\documentclass[letterpaper, 10 pt, conference]{ieeeconf}  

\IEEEoverridecommandlockouts                              

\usepackage{graphicx}
\usepackage{amsmath}
\usepackage{amsfonts}
\usepackage{amssymb}
\usepackage{gensymb}
\usepackage{booktabs}
\RequirePackage[colorlinks=true, allcolors=blue]{hyperref}
\usepackage{orcidlink}  

\usepackage[
    backend=biber,
    style=ieee,
    natbib=true,
    doi=false,
    url=false,
    isbn=false,
]{biblatex}

\AtEveryBibitem{%
   \clearfield{Title}%
   \clearfield{eprint}%
   \clearfield{note}%
}

\let\oldtable\table
\let\endoldtable\endtable
\renewenvironment{table}
  {\oldtable\scriptsize}
  {\endoldtable}

\renewcommand{\citet}[1]{\textcite{#1}}
\renewcommand{\citep}[1]{\parencite{#1}}

\newcommand{\printglossaries}{}

\title{Differentiable Biomechanics Unlocks Opportunities for Markerless Motion Capture}

\author{
R. James Cotton \orcidlink{0000-0001-5714-1400} $^{{1, 2}}$ 
\thanks{
This work was supported by the Research Accelerator Program of the Shirley Ryan AbilityLab and the Restore Center P2C (NIH P2CHD101913). We also thank the Myosuite team for providing early access to the MyoSkeleton model and for being a fun group of people to talk to! Additional thanks goes out to the team from the CottonLab for their help in collecting the data and to all the participants.
}
\thanks{
$^{{1}}$ {Shirley Ryan AbilityLab}
} 
\thanks{
$^{{2}}$ {Northwestern University}
} 
}

\begin{document}

\maketitle


\begin{abstract}
Recent developments have created differentiable physics simulators designed for machine learning pipelines that can be accelerated on a GPU. While these can simulate biomechanical models, these opportunities have not been exploited for biomechanics research or markerless motion capture. We show that these simulators can be used to fit inverse kinematics to markerless motion capture data, including scaling the model to fit the anthropomorphic measurements of an individual. This is performed end-to-end with an implicit representation of the movement trajectory, which is propagated through the forward kinematic model to minimize the error from the 3D markers reprojected into the images. The differential optimizer yields other opportunities, such as adding bundle adjustment during trajectory optimization to refine the extrinsic camera parameters or meta-optimization to improve the base model jointly over trajectories from multiple participants. This approach improves the reprojection error from markerless motion capture over prior methods and produces accurate spatial step parameters compared to an instrumented walkway for control and clinical populations.\\
\end{abstract}


\section{Introduction}

Markerless motion capture (MMC), using multiple synchronized cameras to capture and reconstruct movement using computer vision, can greatly expand access to movement analysis in rehabilitation and movement science research and clinical practice. In recent years several open and proprietary systems have emerged and been validated for this purpose \citep{nakano_evaluation_2020, uhlrich_opencap_2022, kanko_concurrent_2021, cherian_freemocap_2024}. However, existing approaches are limited by the lack of biomechanical body models designed for integration into machine learning pipelines. OpenSim is a widely used open-source biomechanical modeling framework \citep{dembia_opensim_2020, seth_opensim_2018}, for which there are many models including for the whole body and focused models of the arm \citep{rajagopal_full-body_2016, hamner_muscle_2010, mcfarland_musculoskeletal_2023}. However, simulation in this environment is slow and not easily integrated into GPU-accelerated frameworks. The NimblePhysics simulator is faster and supports biomechanical models, but is still not easily integrated into the GPU \citep{nimblephysics}.

Mujoco \citep{todorov_mujoco_2012} is another simulator that supports biomechanical models, including muscle-actuated models. Recently, GPU acceleration was integrated into Mujoco as Mujoco-MJX, based on the physics simulator from Brax \citep{freeman_brax_2021}, although it does not yet support accelerating muscle actuated models. Mujoco has enabled the development of multiple biomechanically oriented frameworks designed integration with machine learning and reinforcement learning, including MyoSuite \citep{caggiano_myosuite_2022, wang_myosim_2022}, which includes a tool for converting OpenSim models into Mujoco, and LocoMujoco \citep{al-hafez_locomujoco_2023}, a locomotion imitation benchmark suite.

In addition to performance, a benefit of simulators like Brax and Mujoco-MJX is that they are differentiable, which opens up the opportunities that we explore in this work. We previously showed that implicit representations for movement trajectories, where a function is learned that maps for time to a coordinate space, $f_\theta: t \rightarrow \mathbf x$,  recover smoother and more anatomically consistent trajectories from markerless motion capture while having better convergence properties \citep{cotton_improved_2023}. To recover joint angle trajectories, we then separately fit inverse kinematics to these trajectories using a biomechanical model in \citep{cotton_markerless_2023}. This was done using a bilevel optimization approach that jointly estimates the skeleton scaling and the inverse kinematics across multiple trajectories \citep{werling_rapid_2022}. We were forced into this two-stage approach because we could not propagate gradients from the reprojected marker location errors through the kinematic model and into the learned implicit representation, which is straightforward to do with differentiable, GPU-accelerated simulators.

However, a barrier to this is that, to the best of our knowledge, scaling the models to fit the anthropomorphic measurements of an individual is not supported with current platforms using Mujoco. Furthermore, there are many different human pose estimation (HPE) algorithms for detecting keypoints from images, that are trained on a diversity of datasets, each having different idiosyncracies. To obtain accurate fits, the model must have marker locations that are optimized for the keypoint set used. Differentiable biomechanical models allow the marker locations on the base model to be optimized end-to-end with the reconstructions from a population of individuals with a range of anthropomorphic measurements. Differentiability also allows marker offsets and skeleton scaling for each individual to integrate naturally into the optimization process. In this work, we selected the MOVI keypoints outputs from the MetrABs-ACAE algorithm \citep{sarandi_learning_2023}, which our prior work found are some of the most accurate and anatomically consistent when varying viewpoints. Full differentiability also means that we can perform bundle adjustment \citep{hartley_multiple_2003} by computing the derivatives with respect to the extrinsic camera parameters. Thus the biomechanically constrained observations of the movement can help improve the calibration.

In short, our contributions are:

\begin{itemize}
\item We enable personalized model scaling and inverse kinematic fitting with bilevel optimization using a GPU-accelerated, fully differentiable biomechanical model using Mujoco.
\item Implement this using an end-to-end differentiable bilevel optimization with an implicit representation of pose trajectories propagated through the forward kinematic biomechanical model to minimize the error from the 3D markers reprojected into the images.
\item Perform bundle adjustment to refine the extrinsic camera parameters during this optimization.
\item Implement a trilevel optimization process that meta-optimizes the base biomechanical model marker locations over a dataset of trajectories from multiple participants.
\end{itemize}

\section{Methods}

\subsection{Participants}

This study was approved by the Northwestern University Institutional Review Board. Participants performed several activities, with the majority performing multiple walking trials, often at different speeds or using different assistive devices. These were performed while participants walked along an instrumented walkway that recorded foot contact and toe-off timing and locations \citep{mcdonough_validity_2001, bilney_concurrent_2003}. Some participants performed additional tasks like the timed-up-and-go and the four-square-step-test, that were included during the fitting process. Our dataset contains a mixed population with able-bodied controls and participants with a history of neurologic conditions, predominantly stroke but some traumatic brain injury, spinal cord injury, peripheral neuropathy, as well as lower limb prosthesis users. There are also several participants with a history of knee osteoarthritis and pediatric participants. For quantifying and comparing performance, we focus on the set of trials that were fit using our previously described biomechanical pipeline \citep{cotton_markerless_2023, cotton_improved_2023}.

\subsection{Markerless Motion Capture}

Multicamera data was collected with a custom system we developed using 8 to 12 FLIR BlackFly S GigE cameras with F1.4/6mm lens \citep{cotton_improved_2023, cotton_markerless_2023}. This system acquires 2048 $\times$ 1536 images at 30 fps that are synchronized using the IEEE1558. Extrinsic and intrinsic calibration was performed using the anipose library \citep{karashchuk_anipose_2020} using a checkerboard. These fitted parameters produce a function for each camera $i$ that projects points in the 3D space of the acquisition volume onto the 2D image plane of each camera:

\begin{equation}
\Pi_i: \mathbf x \rightarrow y, \, \mathbf x \in \mathbb R^3, \, y \in \mathbb R^2
\end{equation}

We used the previously described user interface to select the person of interest from an initial scene reconstruction, which is used to compute the bounding box for that individual across all of the cameras and frames. Keypoints were estimated independently for each frame using an algorithm trained on numerous 3D datasets that outputs the superset of all these formats (MeTRAbs-ACAE) \citep{sarandi_learning_2023}. We used a subset from this output that are the 87 keypoints from the MOVI dataset \citep{ghorbani_movi_2021}, which correspond to those often used in biomechanical analysis and that we used in our prior reconstruction methods \citep{cotton_markerless_2023, cotton_improved_2023}. MeTRAbs-ACAE also directly outputs 3D joint locations from each camera, but we only use the 2D outputs. In our prior work, we have found the keypoints from this algorithm are more geometrically consistent when varying perspectives than other algorithms \citep{cotton_markerless_2023}, likely due to the unique training on extensive 3D datasets. It does not provide confidence intervals for each keypoints, so we approximated them by measuring the standard deviation of each 3D joint location estimated from 10 different augmented versions of each video frame. This was converted to a confidence estimate using a sigmoid function with a half maximum at 30mm and a width of 10mm. This is different than the value we previously used (200 and 50 respectively) \citep{cotton_improved_2023}, as we discovered these prior values provided more of a binary ``detected'' signal than a continuous score.

\subsection{Implicit representation}

When reconstructing biomechanics from multiple synchronized cameras, we previously found an implicit representation that uses a function to map from time to virtual marker positions, $f_\phi(t) \rightarrow \mathbf x$ with $\mathbf x \in J \times 3$, outperforms optimizing an explicit trajectory for $\mathbf x(t)$ or using a robust triangulation \citep{cotton_improved_2023}. This learnable function is instantiated as a multi-layer perceptron (MLP) and time is input using sinusoidal positional encoding after scaling the time inputs from 0 to $\pi$ to prevent aliasing in the encoding \citep{vaswani_attention_2017}. In this work, we modified the implicit representation from directly outputting 3D joint locations to outputting pose parameters that are input to a forward kinematics model implemented in Mujoco. The pose parameters corresponding to joint angles were constrained to the joint range limits through a \texttt{tanh} nonlinearity.

\subsection{Mujoco model}

We used a biomechanically grounded, forward kinematic implemented in Mujoco \citep{todorov_mujoco_2012}. We based our model on the torque-driven model implemented in Mujoco from LocoMujoco \citep{al-hafez_locomujoco_2023}, which in turn is based on the OpenSim model \citet{hamner_muscle_2010}. This torque-driven model can be run with the GPU-accelerated engine for Mujoco, MJX. MJX does not currently support muscle-driven models, but this is under development. We modified the LocoMujoco model by removing collisions other than on the feet, which significantly accelerated the forward kinematics, and we also added a neck joint with 3 degrees of freedom. In our prior work, we optimized marker locations compatible with the MOVI keypoints on a similar biomechanical model in the OpenSim format using the nimblephysics simulator \citep{nimblephysics}. We ported these marker locations to the model used in this work using the site Mujoco element, which corresponds to a virtual marker rigidly attached to a component of the model.

To the best of our knowledge, model scaling, and marker offset calculation have not been implemented using Mujoco and MJX. LocoMujoco implements a similar process by modifying the model specification in the XML before instantiating it, but this does not allow these parameters to be changed dynamically in the GPU or to compute the derivatives or trajectories with respect to scaling and offsets. We implemented a wrapper for the forward kinematics that allows passing in scaling parameters as well as marker offsets in addition to the joint angles. The base model is then modified in memory before computing the forward kinematics. The scaling was implemented using a matrix that maps from a set of scale parameters to scaling individual body segments isotropically. We included 8 scale parameters: an overall size, the pelvis, left thigh, left leg and foot, right thigh, right leg and foot, the left arm, and the left leg. This could be naturally extended to allow separate scales on more body segments or anisotropic scaling, but we did not explore that in this work. Following notation used by the popular skinned multi-person linear model (SMPL)\citep{loper_smpl:_2015}, we can thus represent our GPU-accelerated, biomechanical forward kinematic equation as:

\begin{equation}
{\mathbf x} = \mathcal M (\theta, \beta)
\end{equation}

Where $\theta \in \mathbb R^{40}$ are the inputs to the skeleton model, $\beta \in \mathbb R^{8+87*3}$ are the set of 8 scaling parameters are the offsets for each of the marker locations, and $\mathcal x \in \mathbb R^{87 \times 3}$ are the predicted marker locations. The forward kinematic model also can provide the location and orientation of each of the 21 body components along the kinematic tree as well as the mesh geometry locations and orientations, neither of which we use in this work. Finally, the model includes dynamics, with each component having linear and rotational velocities that respond to torques from the joints, which we also do not use in this work.

\subsection{MyoSkeleton model}

We were also kindly granted early access to a new model being developed for MyoSuite \citep{caggiano_myosuite_2022, wang_myosim_2022}, called the MyoSkeleton. This has many more degrees of freedom (152), such as all the degrees of freedom in the spine at many vertebral levels, a fully articulated hand, and even an articulated patella. These additional degrees of freedom are governed by 66 equality constraints corresponding to typical movement, such as how the flexion at L4-5 should be related to the flexion at L3-4. MJX allows computing an error function for how much these constraints are violated \citep{todorov_mujoco_2012}, $\boldsymbol \epsilon = \mathrm{err}(\boldsymbol \theta)$. When fitting the MyoSkeleton model, in addition to having the implicit function output 152 degrees of freedom, we included an additional loss term that penalized the violation of these constraints,$\mathcal L_{\epsilon} = \lambda_\epsilon \frac{1}{66} \sum_i \epsilon_i^2$. We used $\lambda_\epsilon=0.01$ based on initial experiments showing this drove the constraint violations to very low levels, and leave further optimization of this to future work.

\subsection{Bilevel optimization}

We followed the bilevel optimization approach of \citet{werling_rapid_2022}, jointly solving for the skeleton scaling, marker offsets, and joint angle trajectories over multiple movement sequences. This approach was developed for precomputed 3D marker trajectories, either from traditional optical marker-based motion capture and then adopted in our recent work to virtual 3D marker trajectories from markerless motion capture \citep{cotton_markerless_2023}. In this work, we make two critical modifications: 1) we represent the pose trajectory over time, $\theta_t$, with an implicit representation instead of using this function to predict virtual marker locations, $\mathbf x_t$, and 2) we minimize the error between the detected joint locations in the image plane and the marker locations in 3D space after reprojecting the coordinates through the calibrated camera function, $\Pi_i$,  for camera $i$. Thus, this approach directly optimizes the joint angles that control the biomechanical model to produce marker locations consistent with keypoints detected in the image plane. This is possible because our entire model is end-to-end differentiable and accelerated on the GPU.

More formally, we have a set of $N$ recordings, each of which has keypoints detected for each joint, $j$, per timestep, $t$, and per camera, $c$: $\left\{y^{(n)}_{t,j,c} \right\}_{n=1...N}$. The modeled 3D virtual marker locations are produced from the learned implicit function learned separately for each trajectory, which is then passed through the forward kinematic model using a common $\beta$ parameter for all trajectories:

\begin{equation}
\mathbf x^{(n)}_{t} = \mathcal M(f_{\phi_n}(t), \beta)
\end{equation}

The parameters for the set of trajectories, $\left\{\phi_0, ..., \phi_N, \beta\right\}$, are jointly optimized by minimizing the reprojection loss between the detected 2D keypoints and the reprojected 3D marker locations:

\begin{equation}
\label{eq:reprojection_loss}
\mathcal L_{\Pi} = \frac{1}{T \cdot J \cdot C}\sum_{T, J, c\in C} w_{c,t,j} \, g \left( || \Pi_c \mathbf x_{t,j} - y_{t,j,c} || \right)
\end{equation}

Where $g(\cdot)$ is the Huber loss and $w_{c,t,j}$ is the confidence of the detected keypoint. We also include an L2 regularization on the components of $\beta$ corresponding to the site markers to prevent them from shifting too far from the default location:

\begin{equation}
\label{eq:offset_regularization}
\mathcal L_{\beta} = \frac{1}{87 \times 3} \sum_{i=8...8+87*3} \beta_i^2
\end{equation}

The total loss is then:

\begin{equation}
\mathcal L = \mathcal L_{\Pi} + \lambda_\beta \mathcal L_{\beta} + \lambda_\epsilon \mathcal L_{\epsilon}
\end{equation}

We did not find any additional regularization for temporal smoothness was necessary, although this could be an opportunity for future improvements. We only included the constraint violation loss, $\mathcal L_{\epsilon}$, when fitting the MyoSkeleton model.

\subsubsection{Bundle adjustment}

Because the entire system, including the camera parameters, is end-to-end differentiable, we can also perform bundle adjustment to further refine the calibration parameters. Specifically, we allowed the translation and rotation parameters of $\Pi_c$ from (\ref{eq:reprojection_loss}) to be learnable parameters, while performing the optimization. We used a learning rate schedule that only allowed these parameters to start adjusting later during the optimization process.

\subsubsection{Implementation and Optimization details}

Our implicit representations were implemented in Jax using the Equinox framework \citep{kidger_equinox_2021}. We used an MLP with layer sizes of 128, 256, 512, 1024, 2048, 2048, 4096. Time was scaled to not exceed $\pi$ and was positionally encoded, using an encoding dimension of 29. We performed optimization with the AdamW optimizer from Optax \citep{loshchilov_decoupled_2019, deepmind2020jax}, using $\beta_1=0.8$ and weight decay of $1e -5$. The learning rate used included an exponential decay from an initial value of $1e -4$ to an end value of $1e -7$. We used 40k optimization steps with bundle adjustment enabled at the 30k step using a separate Adam optimizer with a learning rate of $1e -5$. We typically performed this on an A6000 and found it could perform more than 500 iterations a second for a single trajectory of 10 seconds. For typical sessions with 10 trials, this optimization took several minutes. Because Jax operates more efficiently on fixed array sizes (new array sizes trigger tracing and recompiling the code), when training on multiple trajectories with different lengths we would randomly sample a set of 300 time points from each trajectory per iteration.

\subsection{Trilevel optimization}

One challenge is placing all of the markers on the biomechanical model in a location that is consistent with the marker locations produced by the model. When developing our prior method using nimblephysics, we iteratively optimized many separate runs and then applied the mean marker offsets to the original unscaled model. Several rounds of this process did improve our results \citep{cotton_markerless_2023}. However, this process required repeatedly correcting some of the marker locations that shifted too far and did not necessarily produce the optimal locations.

We took advantage of the fact that our framework is fully differentiable and runs extremely quickly to jointly optimize the default marker locations by fitting the model to 31 different individuals from our dataset with 2 trials per individual, which we found could be performed on a single 80GB A100. This optimization included multiple of the 31 independent models similar to the one described under bilevel optimization. However, the marker offset for the base model was also made learnable, with the models from the 31 individuals each being independently scaled $\mathcal L_\beta$ applied to their independent site offsets. To prevent the markers from shrinking inward, which we had initially observed, we froze the position of the bilateral heel markers on the base model. We performed this process once at the beginning of this work and then used this optimized base model to fit all of the trajectories for each individual using a bilevel optimization approach, which then scaled the model and learned individualized marker offsets with regularization from (\ref{eq:offset_regularization}) to prevent the markers moving much beyond this.

\subsection{Baseline biomechanics}

We compare the biomechanics results against our prior methodology, which decomposes this into two steps equivalent to a typical biomechanics pipeline \citep{cotton_improved_2023, cotton_markerless_2023}.  Specifically, it uses an implicit representation to reconstruct the 3D marker trajectories over time, which we found produced better results than optimizing explicit marker trajectories directly or robust triangulation. It then uses the nimblephysics library for bilevel optimization, which fits the model scaling and marker offsets while performing inverse kinematics for each trial. However, in contrast to the approach proposed in this paper, this prior approach is not end-to-end differentiable so the nimblephysics model results are optimized against the reconstructed marker trajectories instead of the raw image observations. We hypothesized this decomposition would limit performance.

\subsection{Performance metrics}

\subsubsection{Geometric Consistency}

We measured the geometric consistency between the model-based marker trajectories and the detected keypoints, based on the distance between the reprojected model-based markers and the detected keypoints. To quantify this, we computed the fraction of the points below a threshold number of pixels, conditioned on being greater than a specified confidence interval.

\begin{equation}
\delta_{t,j,c} = || \Pi_c \mathbf x_{t,j} - y_{t,j,c} ||
\end{equation}

\begin{equation}
q(d,\lambda) = \frac{\sum (\delta_{t,j,c} < d)(w_{t,j,c} > \lambda)}{\sum w_{t,j,c} > \lambda}
\end{equation}

Below, we report $GC_d=q(d, 0.5)$ with $d=5$ pixels.

\subsubsection{GaitRite comparison}

We compared the heel marker position from the forward kinematics to those measured by the GaitRite to measure the error of each step length, stride length, and step width. We report the width of the error distribution as normalized IQR, $\sigma_{IQR}=0.7413 \cdot IQR(x)$. We use this in place of the standard deviation because it is more robust to the few outliers we observe, and we also show the histogram of errors in the results.

\section{Results}

\subsection{Visualizations}

\begin{figure}[!htbp]
\centering
\includegraphics[width=1\linewidth]{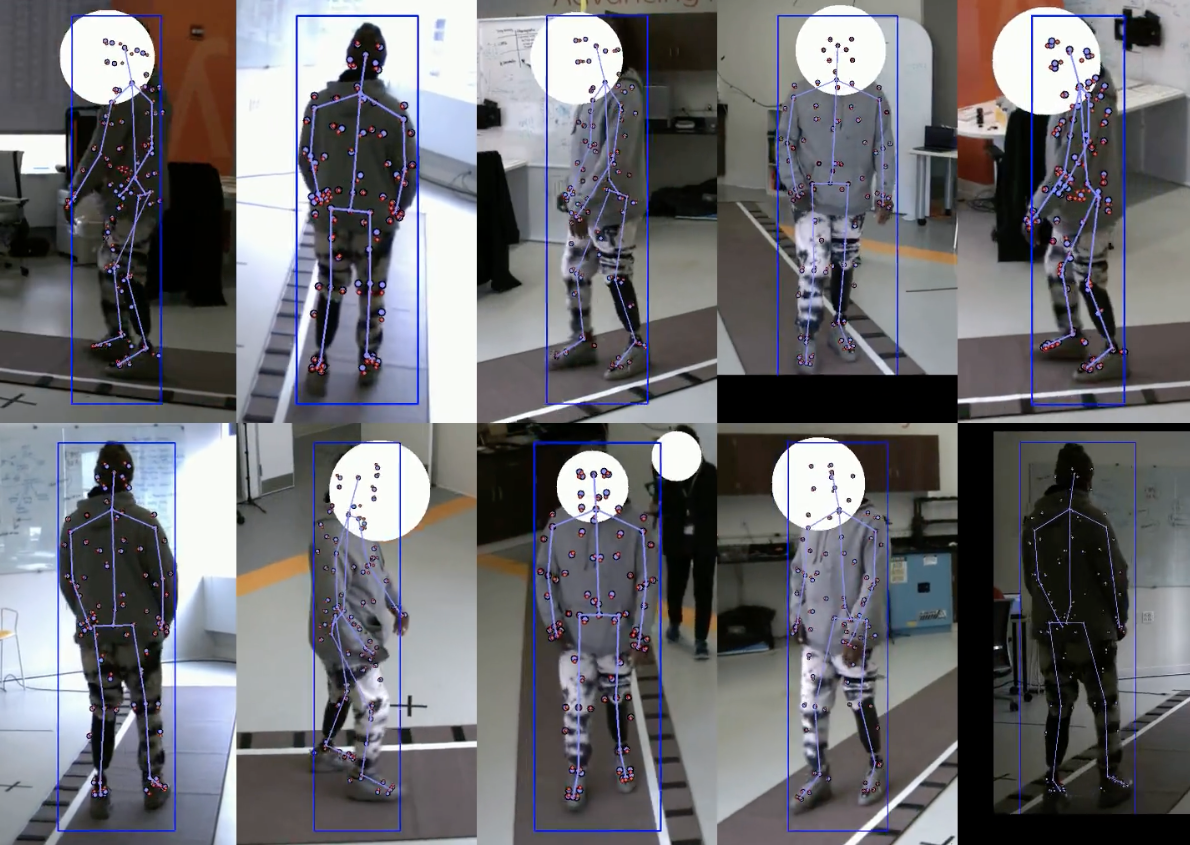}
\caption[]{Visualization of reprojection errors. Each panel is a zoomed in crop from the image around the detected bounding box from a different view. The red dots are the 87 detected keypoints in each frame and the blue dots are the reprojected model-based markers, with the drawn skeleton connecting a subset of these keypoints.}
\label{fig:reprojections}
\end{figure}

We first visually confirmed the quality of the reconstructions by creating videos comparing the reprojected model-based markers reprojected into each of the image planes to the 2D keypoint detections. Figure~\ref{fig:reprojections} shows an example frame from 10 cameras acquired simultaneously, showing close alignment between the reconstructed and detected keypoints.

\begin{figure}[!htbp]
\centering
\includegraphics[width=1\linewidth]{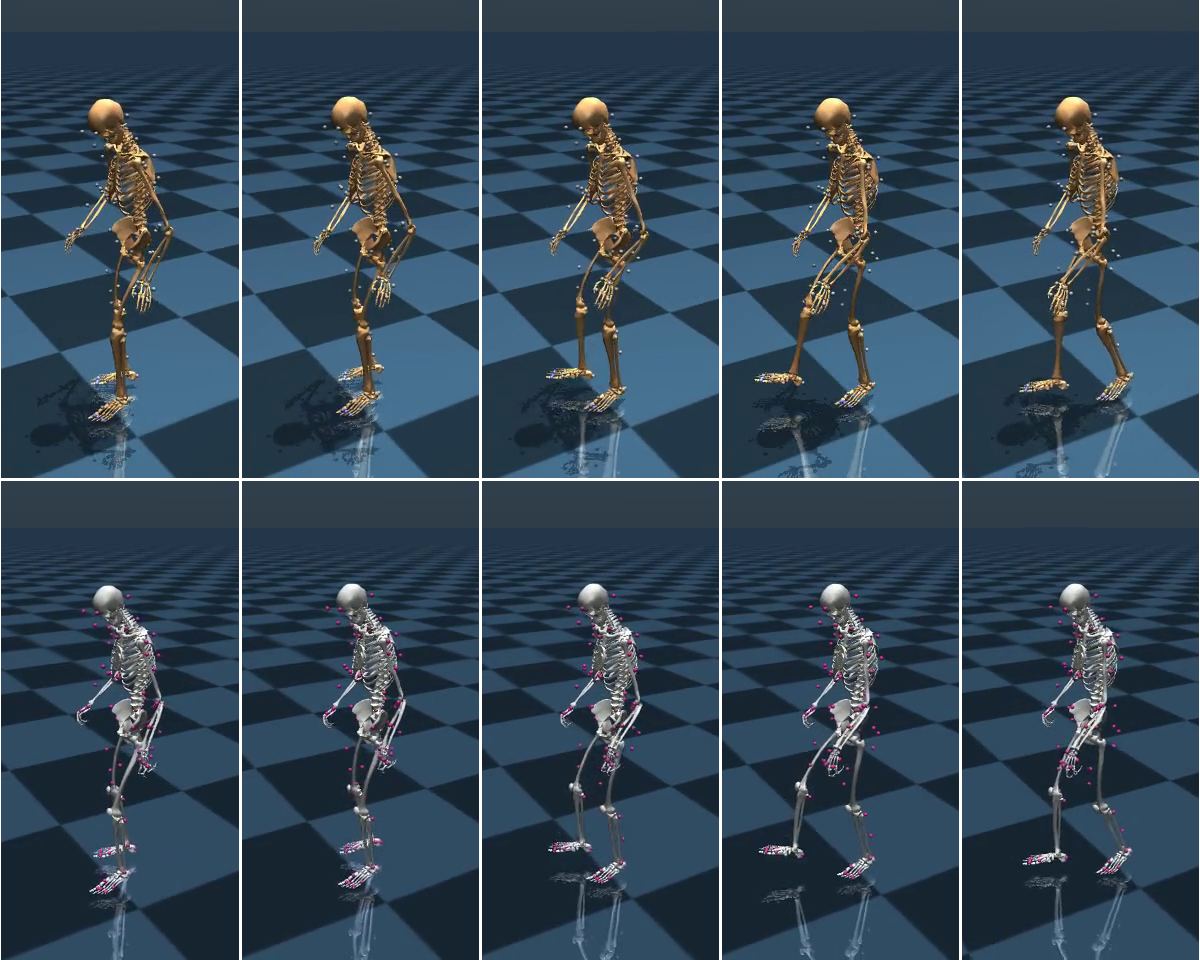}
\caption[]{Sample rendered frames from a walking trial fit from an individual with a history of stroke using a cane on the left side. The top row is the modified LocoMujoco skeleton and the bottom row is the MyoSkeleton.}
\label{fig:walking_frames}
\end{figure}

We also used the Mujoco renderer videos of the walking trajectories, which replicated the movements seen in the videos. Figure~\ref{fig:walking_frames} shows example frames from a walking trial, reconstructed with both skeletons. Note that the fully articulated MyoSkeleton hand is not properly constrained by the 87 keypoints used.

\subsection{Samples waveforms}

\begin{figure}[!htbp]
\centering
\includegraphics[width=0.8\linewidth]{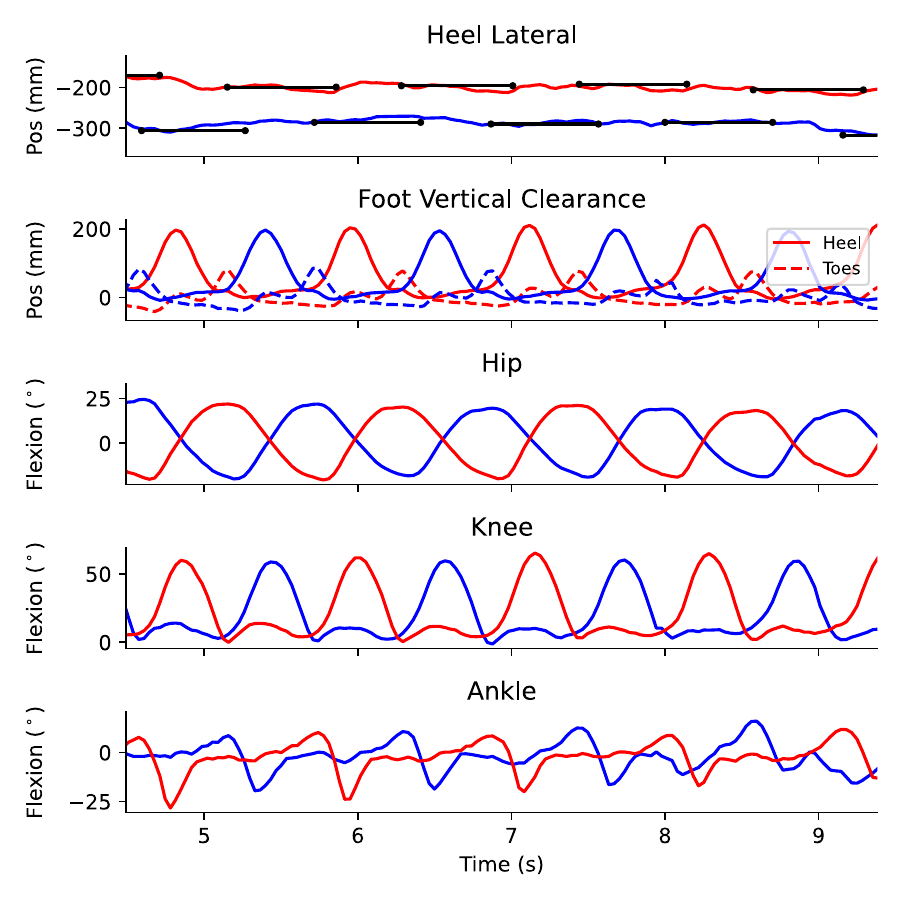}
\caption[]{Sample waveforms from a control participant. The left leg is shown in blue and the right leg is shown in red. The top row shows the lateral position of the heel with the positions detected from the instrumented walkway shown in black. The next row shows the vertical clearance from the heel and toes. The last three rows show the flexion angles for the hip, knee, and ankle, respectively.}
\label{fig:gait_waveforms}
\end{figure}

\begin{figure}[!htbp]
\centering
\includegraphics[width=0.8\linewidth]{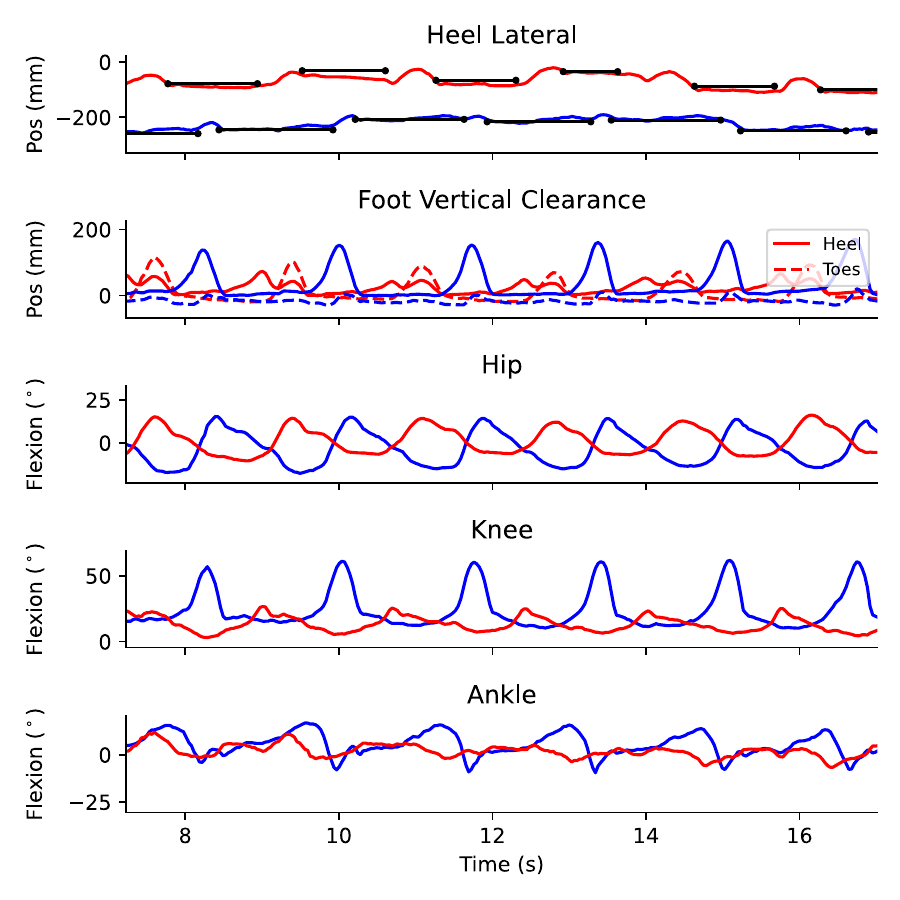}
\caption[]{Sample waveforms from a participant with a history of stroke, corresponding to the visualization from Figure~\ref{fig:walking_frames}. This captures the reduced range of motion on the right hemiparetic side, particularly with reduced knee flexion.}
\label{fig:gait_waveforms_impaired}
\end{figure}

Figure~\ref{fig:gait_waveforms} shows sample waveforms from a single trial from a control participant reflecting a stable, symmetric, periodic gait pattern. Figure~\ref{fig:gait_waveforms_impaired} shows a sample waveform from a participant with a history of stroke which captures the spatiotemporal asymmetry and greater step-width variability.

\subsection{Reprojection loss}

\begin{figure}[!htbp]
\centering
\includegraphics[width=1\linewidth]{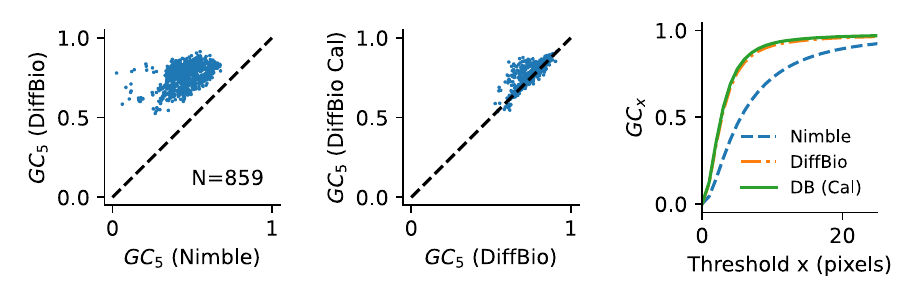}
\caption[]{Reprojection losses for reconstruction with two-stage approach using implicit representation followed by nimblephysics versus end-to-end approach with implicit representation and forward kinematic biomechanical model in mujoco. The left panel compares the $GC_5$  between our prior two-stage approach using nimble physics (horizontal axis) with differentiable biomechanics (vertical axis) for a number of trials. The middle panel compares trials with and without bundle adjustment to refine calibration. The right panel shows the average $GC_x$ as a function of the pixel threshold for those trials.}
\label{fig:reprojection}
\end{figure}

We also compared the error in the camera reprojections between our prior approach and the new approach, measured by the fraction of sites that reproject within a specific number of pixels, $GC_x$. We found that the end-to-end optimization approach with differentiable biomechanics had a better $GC_5$  for almost every trial. When comparing the $GC_5$ for trials using differentiable biomechanics without and with bundle adjustment enabled, we also saw a performance improvement. The average $GC_x$ curve for differentiable biomechanics with calibration exceeded the other approach for all $x$ values (Figure~\ref{fig:reprojection}).

We quantified this difference by performing an ANOVA on as the main factor and $GC_5$ as the dependent variable. This was highly significant ($p < 1 \times 10^{ -5}$, $F = 4418$) and a post-hoc Tukey test showed a significant difference between all conditions. This included the small $1.9\%$ increase in $GC_5$ when enabling bundle adjustment. Note, that this analysis only included 851 of the trials in our dataset, as we do not routinely run reconstruction with Nimblephysics anymore.

\subsection{Step parameter errors}

\begin{figure}[!htbp]
\centering
\includegraphics[width=1\linewidth]{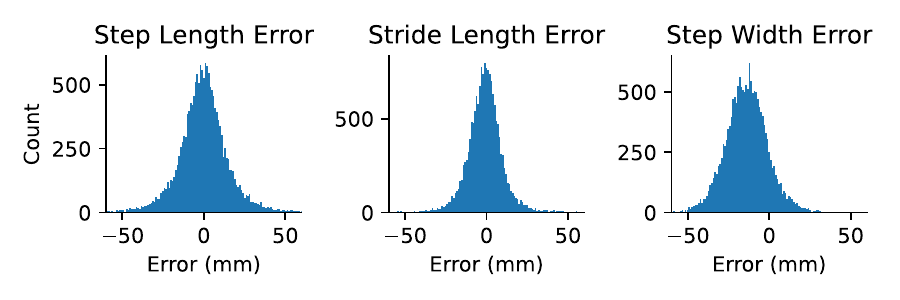}
\caption[]{Histogram of step length, stride length, and step width errors comparing the GaitRite and biomechanical reconstruction.}
\label{fig:step_errors}
\end{figure}

We also compare the errors in the spatial step parameters computed between the GaitRite and the heel marker from the fitted model. The process of aligning the reference frame spatially and temporally between the two systems was the same as our prior work \citep{cotton_improved_2023}, with an affine transformation and temporal offset computed to minimize the error over a set of trials.  Note this marker location does not precisely correspond to the same anatomic location as the center of pressure from the heel measured by the instrumented walkway, but provides a rough estimate of accuracy.  Figure~\ref{fig:step_errors} shows the distribution of errors for each of these parameters.

Table~\ref{table:metrics} shows the $\sigma_{IQR}$ for each of these parameters broken down by several clinical populations. We see for control participants, the error is about 8-10 mm across the parameters but is slightly higher for some of the clinical populations. In particular, for some of the clinical populations, the step length error increases more than the stride length error. We hypothesize this may arise due to asymmetries in weight loading that are measured differently from the instrumented walkway than from the location of heel markers. This may also be due to reduced keypoint detection accuracy of HPE algorithms in some clinical populations, which we have shown for lower limb prosthesis users (LLPUs), particularly those with more proximal or bilateral amputations \citep{cimorelli_portable_2022}. Finally, the error may also arise from keypoints detected on physical therapists who are often walking closely to the participant for safety, which are then misattributed to the participant, although our Huber loss is generally robust to this \citep{cotton_improved_2023}.

\begin{table}
\centering
\caption[]{$\sigma_{\mathrm{NIQR}}$ for gait metrics broken down by clinical population, as well as the number of subjects and trials in each category.}
\label{table:metrics}
\begin{tabular}{p{\dimexpr 0.167\linewidth-2\tabcolsep}p{\dimexpr 0.167\linewidth-2\tabcolsep}p{\dimexpr 0.167\linewidth-2\tabcolsep}p{\dimexpr 0.167\linewidth-2\tabcolsep}p{\dimexpr 0.167\linewidth-2\tabcolsep}p{\dimexpr 0.167\linewidth-2\tabcolsep}}
\toprule
Population & Step Width & Stride Length & Step Length & Subjects & Trials \\
\hline
Controls & 10.6 & 8.1 & 7.9 & 29 & 622 \\
LLPU & 10.6 & 9.0 & 15.1 & 46 & 431 \\
Neurologic & 12.4 & 10.2 & 19.1 & 37 & 333 \\
Orthopedic & 15.3 & 8.5 & 7.9 & 5 & 37 \\
Pediatric & 16.2 & 13.1 & 15.1 & 5 & 34 \\
\bottomrule
\end{tabular}
\end{table}

We also computed this for the LocoMujoco model and the MyoSkeleton model, both with and without bundle adjustment. We found that both models perform well, despite the MyoSkeleton having higher degrees of freedom.

\begin{table}
\centering
\caption[]{Gait metrics over all trials for different models and optimization approaches.}
\label{table:comparison}
\begin{tabular}{p{\dimexpr 0.167\linewidth-2\tabcolsep}p{\dimexpr 0.167\linewidth-2\tabcolsep}p{\dimexpr 0.167\linewidth-2\tabcolsep}p{\dimexpr 0.167\linewidth-2\tabcolsep}p{\dimexpr 0.167\linewidth-2\tabcolsep}p{\dimexpr 0.167\linewidth-2\tabcolsep}}
\toprule
Model & Bund Adj & Step Width & Stride Length & Step Length & $GC_5$ \\
\hline
LocoMujoco & False & 12.31 & 9.97 & 13.00 & 0.70 \\
LocoMujoco & True & 12.19 & 9.05 & 12.57 & 0.72 \\
MyoSkeleton & False & 12.33 & 9.91 & 13.17 & 0.69 \\
MyoSkeleton & True & 12.18 & 9.07 & 12.73 & 0.72 \\
\bottomrule
\end{tabular}
\end{table}

\section{Discussion}

Having a fully differentiable biomechanical model that can be accelerated on a GPU is an extremely powerful tool for rehabilitation and movement science research. Here, we showed we can drive this forward kinematic with implicit representation for movement trajectories to get accurate movement reconstructions. The trajectories from multiple trials can be jointly optimized with the skeleton scaling and marker offsets in a consistent end-to-end framework, providing a straightforward implementation of bilevel optimization  \citep{werling_rapid_2022}.

Our model also allows trilevel optimization, where multiple individuals' trajectories and skeleton scaling are fit simultaneously while also optimizing the base model marker offsets. This is a powerful tool for integrating biomechanics into different computer-vision pipelines, as in our experience, inconsistencies between marker conventions are a source of error. Now, these markers can be roughly initialized and attached to the right body component and then fine-tuned on an entire dataset. This will be convenient for creating new models for new HPE algorithms, such as detailed hand tracking or other commonly used markersets. We did note some limitations from our trilevel approach. The first limitation was that only 31 individuals with 2 trials each were used due to limitations in GPU memory, and this many people might not be sufficiently representative. This limitation arose as all trials were processed for each iteration. In the future, we will implement mini-batching over subsets to allow jointly optimizing parameters over many participants. We are also experimenting with manually setting and anchoring additional markers that are anatomically well-defined in addition to the heel, as we saw that the elbow site location was displaced more anterior than we believe it should be. Thus, the current model risks introducing a biased offset in the joint angles compared to ground truth. Replicating the overlay visualizations from our prior work \citep{cotton_markerless_2023} would also be useful to confirm the model is not introducing any systematic errors.

Our results show that end-to-end optimization produces fits that are more geometrically consistent with the detected keypoints than a two-stage optimization approach, as measured through the $GC_x$ of the reprojected fit results. This is fairly unsurprising, as decomposing the problem does not allow the initial trajectories to be directly constrained by the biomechanical model. Our prior work with the implicit representation included explicit regularization to keep the bone lengths consistent over the trajectory for certain keypoints  \citep{cotton_improved_2023}, but for many keypoints, such as those in the foot or the middle of a limb, it was less clear how to structure those constraints. In our fully differentiable implicit reconstruction approach, these constraints and others such as joint limits arise naturally from the biomechanical model fit for each individual over multiple trials. In the case of the MyoSkeleton, this also extended to the joint equality constraints to allow fitting the many degrees of freedom over the spine with anatomically valid values.

We exploited differentiability through the camera model to incorporate bundle adjustment into the later iterations of optimization, to refine extrinsic camera parameters parameters. This step made a small but significant improvement in the reprojection errors and the step parameter errors. This is despite us already performing a rigorous calibration with anipose before each experiment \citep{karashchuk_anipose_2020}. While we are not ready to skip the calibration step entirely, ultimately removing this need would be another useful step to making MMC even easier to use. This also suggests a path to optimizing other differentiable meta-parameters for a session, such as how we account for the keypoint confidence scores.

Optimizing how we account for keypoint uncertainty is another opportunity to improve the reconstructions we have not yet explored. Critically, our current approach does not produce confidence bounds on the joint angles, although these will vary across a trial based on the number of cameras viewing certain markers and their relative geometry. Reliable confidence estimates are critical for wider usage, particularly in clinical settings, as knowing when you can trust the data is essential. In the context of lifting from 2D to 3D keypoints, we have shown that fitting probabilistic models and optimizing the negative log-likelihood (NLL) instead of minimizing a point estimate of error can produce these calibrated uncertainty estimates \citep{pierzchlewicz_optimizing_2023}. A promising future direction is extending the trajectory representation to a probabilistic one and minimizing the NLL, to obtain these confidences for MCC. This will allow us to account for the noise in the keypoints and their internal consistency and produce calibrated confidence intervals around our joint angle estimates. It will also require us to develop validated metrics-of-merit that assess the internal consistency of model fits and uncertainty estimates, as frequently ground truth is not available. We anticipate that the trilevel optimization of models fit to multiple individuals simultaneously will allow us to meta-optimize some of these parameters very efficiently.

Computing confidence of joint angles will also be beneficial for approaches were obtaining ground truth information is hard. This includes large scale data collection on representative clinical populations, as in our current work. This also includes tracking the hand, where placing markers on individual joints is impractical and alters the visual appearance and function of the hand. The MyoSkeleton \citep{caggiano_myosuite_2022, wang_myosim_2022} includes a fully articulated hand, and we have also ported the MoBL-ARMS biomechanical model \citep{holzbaur_model_2005}. Our preliminary results show this end-to-end model works well for tracking individual fingers.

It is reassuring that our model produces high-quality fits for a range of clinical populations, including lower-limb prosthesis users, those with neurologic conditions, pediatric patients, and people with musculoskeletal conditions. The step parameters do increase in some populations, with the noise in the step length increasing to as high as 2cm for individuals with neurologic conditions who are often walking with an accompanying therapist, where keypoints from the wrong individual may be mis-associated. This suggests opportunities for improving the geometric scene understanding and using multi-person tracking to further improve the results. Stride length does remain at 1cm in this condition, which may also suggest this relates to systematic differences between measuring weight pressure locations versus computer vision. Explicitly modeling uncertainty could further help in these cases.

Beyond these improvements for fitting biomechanical models to MMC data, we are more broadly excited about having a GPU-accelerated biomechanical model used consistently across different lines of research. For example, we recently showed that using reinforcement learning (RL) to train an imitation learning policy driven by monocular computer vision improves gait analysis from single camera views \citep{smyrnakis_advancing_2024}, but this work would benefit from replacing the simple humanoid model with the model we describe here. Recent work shows that an RL policy imitation learned on a large-scale motion dataset can be distilled for generative modeling with motions conditioned on learned latent codes \citep{luo_universal_2023}. This could be combined with our work on self-supervised learning of gait-based biomarkers \citep{cotton_self-supervised_2023} to learn latent codes (i.e., biomarkers) that characterize certain gait patterns and then generate motions representative of those codes, or possibly even predict motions on different tasks. Finally, MyoSuite is building a community and set of tools around neuromuscular and biomechanical modeling using Mujoco, and the work we described here will enable the creation of large-scale datasets of movements from real people compatible with their tools \citep{caggiano_myosuite_2022}. This hybridizes well with recent advances such as the SKEL and OSSO model \citep{keller_skin_2023, keller_osso_2022}, which map biomechanical models to the body models more typically used in computer vision research, such as SMPL \citep{loper_smpl:_2015}.

\section{Conclusion}

Fully differentiable biomechanical models that can be accelerated on a GPU are a promising tool for rehabilitation and movement science. Here we showed this naturally allows us to implement an end-to-end bilevel optimization algorithm to scale biomechanical models and fit inverse kinematics to multiple trials while directly minimizing the errors reprojected back into the camera planes. It even allows performing bundle adjustment to refine camera extrinsic parameters and trilevel optimization to meta-optimize a model over a set of individuals. This work will enable a consistent biomechanically-ground model to be used across different lines of research hybridizing between computer vision, machine learning, and biomechanics.
\printglossaries




{\small
\printbibliography
}

\end{document}